# Automated Parsing of Engineering Drawings for Structured Information Extraction Using a Fine-tuned Document Understanding Transformer

Muhammad Tayyab Khan[1,3#], Zane Yong[2#], Lequn Chen[2], Jun Ming Tan[2], Wenhe Feng[1], and Seung Ki Moon[3*]

[1] Singapore Institute of Manufacturing Technology (SIMTech), Agency for Science, Technology and Research (A*STAR), 5 CleanTech Loop, #01-01 CleanTech Two Block B, Singapore 636732, Republic of Singapore

[2] Advanced Remanufacturing and Technology Centre (ARTC), Agency for Science, Technology and Research (A*STAR), 3 CleanTech Loop, #01-01 CleanTech Two, Singapore 637143, Republic of Singapore

[3] School of Mechanical and Aerospace Engineering, Nanyang Technological University, 639798, Singapore

[*] skmoon@ntu.edu.sg (S.K. Moon) [#] Equal contribution

*Abstract* - **Accurate extraction of key information from 2D engineering drawings is crucial for high-precision manufacturing. Manual extraction is slow and labor-intensive, while traditional Optical Character Recognition (OCR) techniques often struggle with complex layouts and overlapping symbols, resulting in unstructured outputs. To address these challenges, this paper proposes a novel hybrid deep learning framework for structured information extraction by integrating an Oriented Bounding Box (OBB) detection model with a transformer-based document parsing model (Donut). An in-house annotated dataset is used to train YOLOv11 for detecting nine key categories: Geometric Dimensioning and Tolerancing (GD&T), General Tolerances, Measures, Materials, Notes, Radii, Surface Roughness, Threads, and Title Blocks. Detected OBBs are cropped into images and labeled to fine-tune Donut for structured JSON output. Fine-tuning strategies include a single model trained across all categories and category-specific models. Results show that the single model consistently outperforms category-specific ones across all evaluation metrics, achieving higher precision (94.77% for GD&T), recall (100% for most categories), and F1 score (97.3%), while reducing hallucinations (5.23%). The proposed framework improves accuracy, reduces manual effort, and supports scalable deployment in precision-driven industries.**

*Keywords* - **2D Engineering Drawings, Deep Learning, Document Understanding Transformer (Donut), Fine-Tuning, Structured Information Extraction, YOLOv11**

## I. INTRODUCTION

Accurate extraction of critical information from 2D engineering drawings is essential for precision manufacturing, directly impacting product quality, efficiency, and cost-effectiveness [1], [2], [3]. These drawings contain technical specifications that guide key decisions, such as manufacturing tool and process selection, which ultimately affect production costs and timelines. Traditionally, this information is extracted manually by experienced engineers who interpret annotations on the drawings [4]. However, manual extraction is time-consuming, error-prone, and especially challenging when dealing with intricate symbols or densely annotated layouts. Misinterpretations can lead to incorrect tooling and machining parameter settings, defective components, and significant production delays [5].

Existing semi-automated approaches, such as ballooning techniques [6] and commercial tools like Mitutoyo's MeasurLink [7], offer partial automation but still rely heavily on human input. This reliance limits scalability and introduces inconsistencies. Machine learning-based methods, including object detection using YOLO [8] and Optical Character Recognition (OCR) tools [9], have shown promise but face notable challenges. Generic OCR models often produce segmentation errors when handling overlapping annotations, non-standard symbols, or complex layouts. These limitations largely arise from a lack of domain-specific fine-tuning, as most OCR systems are trained on general datasets rather than engineering documents.

Khan et al. [5] proposed a fine-tuned OCR-based method tailored to 2D drawings, which improved accuracy. However, it was designed to extract all Geometric Dimensioning and Tolerancing (GD&T) annotations from entire drawings, and its performance degraded with more complex layouts. This highlights the need for a more robust segmentation strategy. A YOLO-based Oriented Bounding Box (OBB) detection model can isolate relevant regions, simplifying subsequent processing steps.

To address these challenges, this paper presents a novel hybrid deep learning framework for structured information parsing by integrating YOLOv11-obb for annotation region detection with a transformer-based Donut model. The system generates structured JSON outputs using an OCR-free approach. Two training strategies, a single model and a category-specific model, are explored to evaluate the trade-offs between generalization and precision. The framework is designed to reduce manual effort and enable scalable, accurate information extraction for precision-driven manufacturing applications.

## II. METHODOLOGY

This study adopts a two-stage deep learning framework to extract structured information from 2D engineering drawings. The system integrates YOLOv11 for OBB detection and a transformer-based Donut model to parse the content within the detected regions, as shown in Fig. 1.

A total of 1,367 2D engineering drawings are collected from publicly available sources. These drawings contain annotations across nine key categories: GD&Ts, General Tolerances, Materials, Measures, Notes, Radii, Surface Roughness, Threads, and Title Blocks. Manual annotation is performed using the Computer Vision Annotation Tool (CVAT) [10], with each instance labeled according to its corresponding category.

YOLOv11 is trained on this annotated dataset to detect OBBs corresponding to the nine feature types. From the detected regions, 11,469 image patches are extracted, each containing a single annotated feature. These patches vary in layout complexity, font style, and annotation density. To reduce the burden of exhaustive manual labeling, a representative subset of 1,000 patches is manually annotated by domain experts. This subset is evenly distributed across all nine categories to ensure balance and diversity for downstream model fine-tuning. This approach minimizes annotation effort while preserving full category coverage.

To improve model generalization, the labeled dataset is expanded through data augmentation using the PyTorch library [11]. These transformations simulate real-world variability commonly found in engineering documents, such as scanning artifacts, inconsistent print quality, and font distortions. The following augmentation techniques are applied:

1. Sharpness adjustment by randomly selecting between blurred, original, or sharpened versions.
2. Contrast modification with a probability of 0.5.
3. Rotation by 0, 90, 180, or 270 degrees.
4. Grayscale conversion with a probability of 0.5.
5. Image inversion with a probability of 0.5.

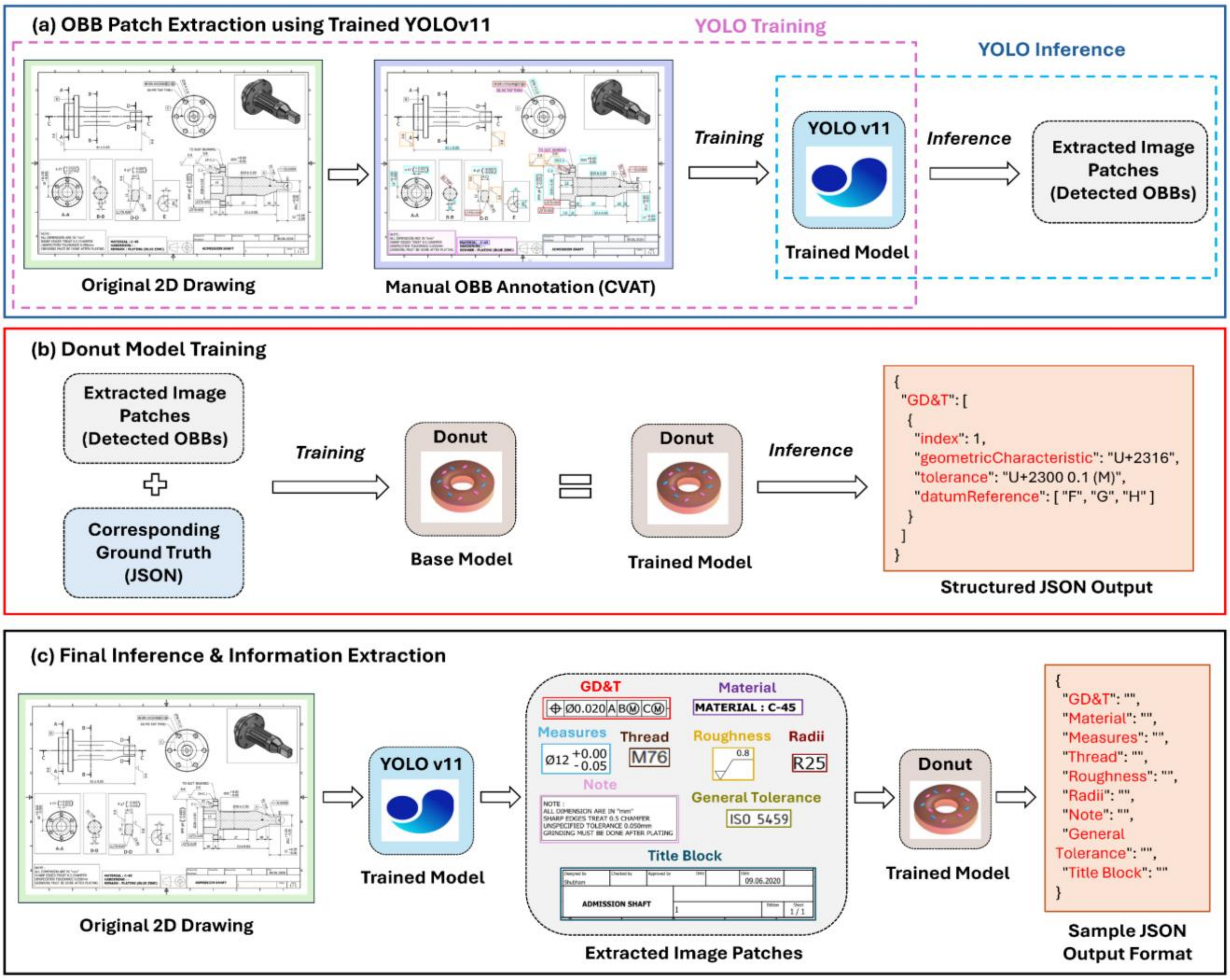

Fig. 1. Overview of the proposed hybrid deep learning framework for structured information extraction from 2D engineering drawings. (a) YOLOv11 is trained to detect OBBs across annotated 2D drawings and extract relevant image patches. (b) Donut is fine-tuned using the extracted image patches and corresponding JSON ground truth to learn structured annotation parsing. (c) During inference, YOLOv11 detects annotation regions from unseen drawings, which are parsed by the trained Donut model to generate structured outputs in JSON format.

These augmentations expand the training dataset from 1,000 to 6,000 image-label pairs, enhancing visual diversity and improving robustness during model training, as shown in Fig. 2.

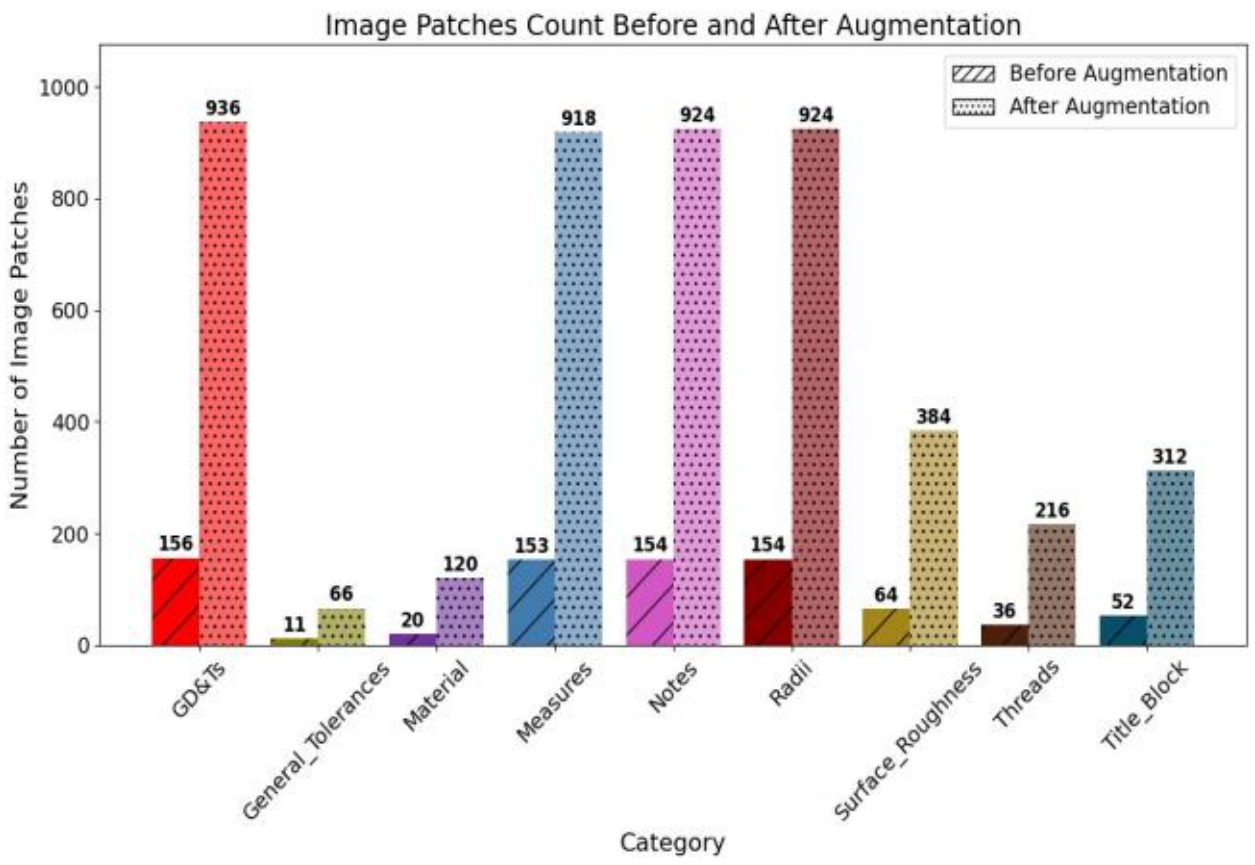

Fig. 2. Distribution of image patches across nine feature categories before and after data augmentation.

A visual example of an original patch, its augmented variants, and the corresponding structured JSON ground truth used for Donut training is shown in Fig. 3, demonstrating the format and diversity of the fine-tuning dataset.

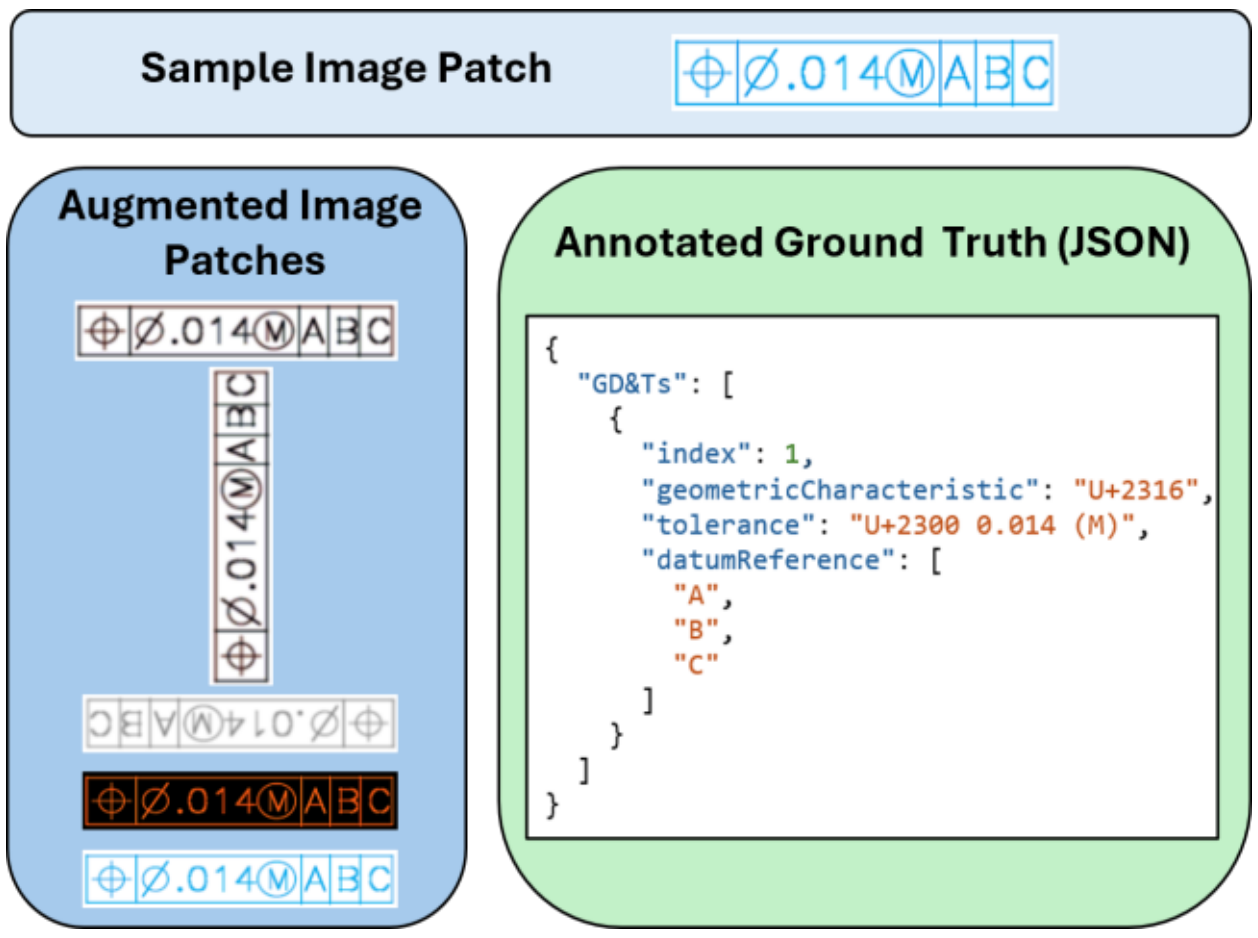

Fig. 3. A sample image patch with augmented variants and corresponding structured JSON ground truth used for Donut fine-tuning.

Following OBB detection, Donut is fine-tuned on the cropped image patches to extract structured textual and geometric information in JSON format. Its transformer-based architecture processes visual and textual content jointly, eliminating the need for traditional OCR or explicit text segmentation. This makes it particularly effective for parsing 2D drawings with overlapping, stylized, or irregular annotations.

Two training strategies are explored. The first strategy is a single model approach, where a single Donut model is trained across all nine feature categories. The second strategy is a category-specific approach, where nine individual Donut models are trained for each category type. The models are fine-tuned using cross-entropy loss and the Adam optimizer with a batch size of 1 for 30 epochs. All models are trained on a single NVIDIA GeForce RTX 4090 GPU with identical hyperparameters. Training the single model requires approximately 20 hours, whereas individual category-specific models train substantially faster, ranging from several minutes to a few hours depending on the respective dataset sizes.

Model performance is evaluated using four key metrics: precision, recall, F1 score, and hallucination rate. The precision measures the proportion of correct predictions among all outputs, while the recall captures the proportion of actual features successfully identified. The F1 score provides a balanced measure of both precision and recall. The hallucination rate quantifies how frequently the model incorrectly predicts features that do not exist in the input. This evaluation framework provides a clear basis for comparing generalization capability versus category-specific accuracy and supports identification of the optimal strategy for information parsing in drawings.

## III. RESULTS

The performance of the single model and category-specific training strategies across nine annotation categories is evaluated using four metrics: precision, recall, F1 score, and hallucination rate, as summarized in Fig. 4.

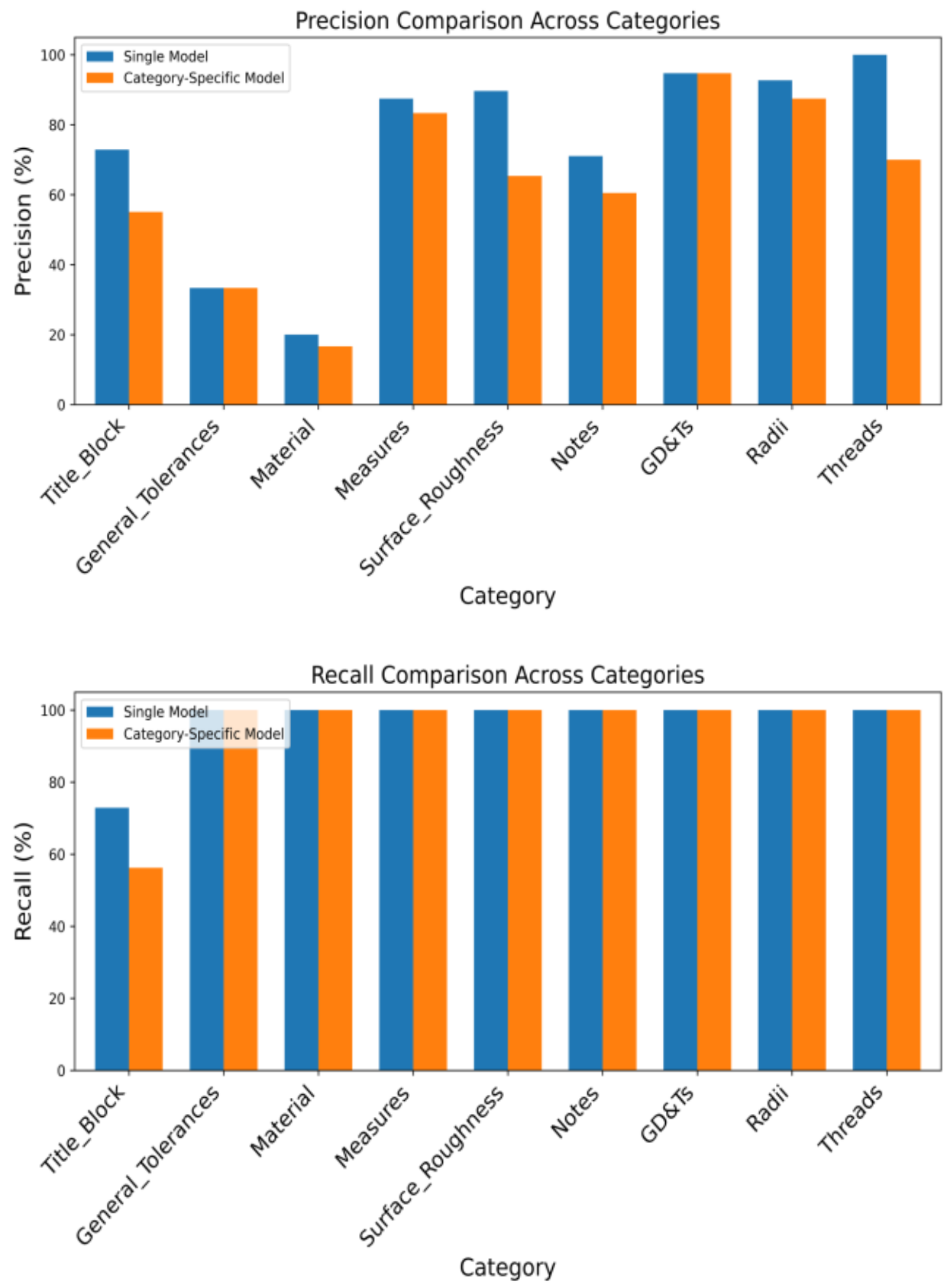

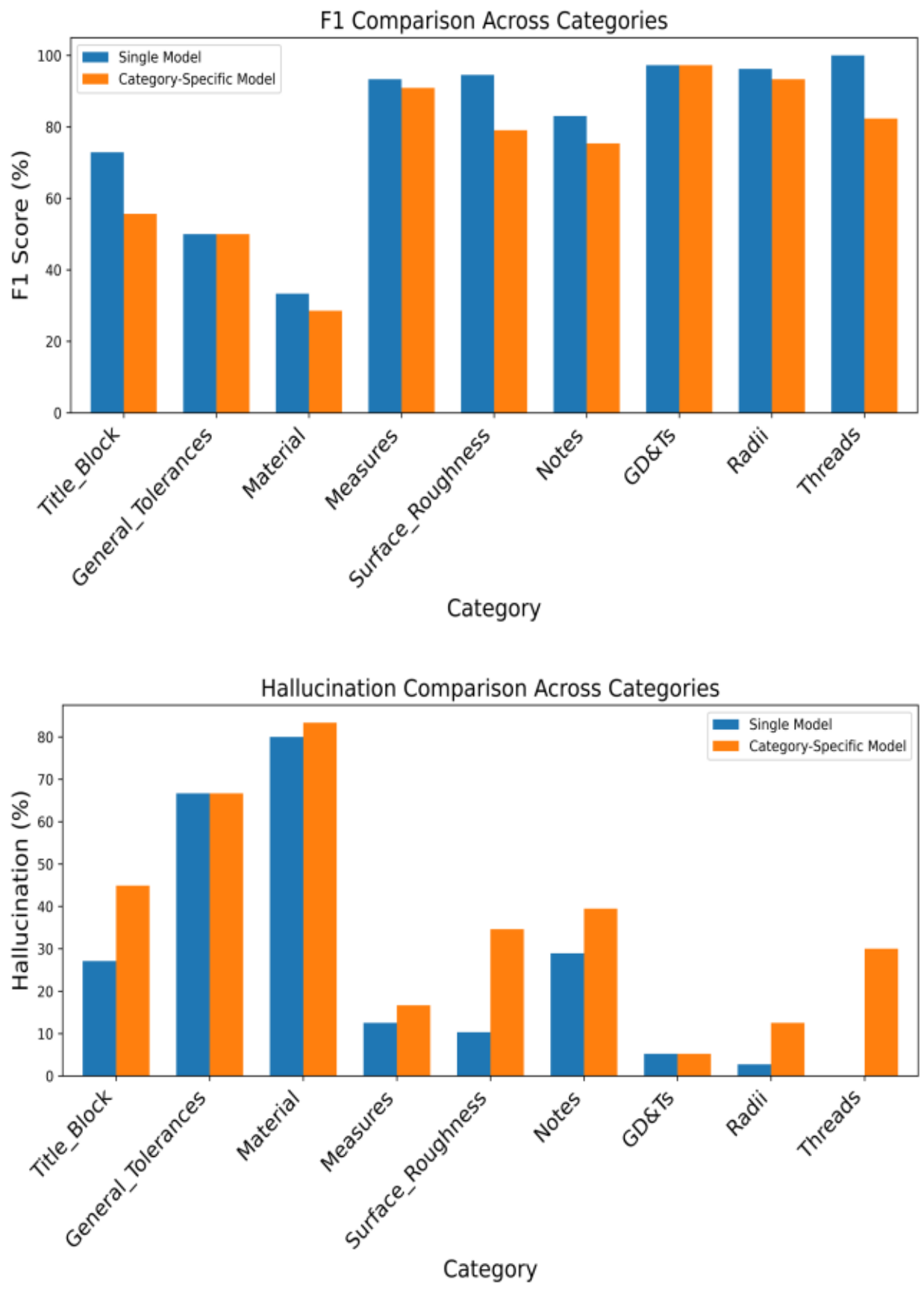

Fig. 4. Performance comparison between single model and category-specific training strategies across nine feature categories using four evaluation metrics: (a) Precision, (b) Recall, (c) F1 score, and (d) Hallucination rate.

The single model consistently achieves higher precision across all feature categories. For instance, the precision for *Surface Roughness* is 89.7% in the single model compared to 65.4% in the category-specific models. Similarly, precision improves for *Notes* (from 60.5% to 71.1%), *Threads* (from 70% to 100%), *Title Blocks* (from 55.1% to 72.9%), and *Radii* (from 87.5% to 92.7%). This consistent advantage suggests that the single model is more effective at minimizing false positives across diverse categories, likely due to its exposure to the full annotation space and broader contextual patterns during training.

Recall remains consistently high across almost all feature categories for both models. Most feature types achieve perfect recall (100%) for *General Tolerances*, *Materials*, *Measures*, *Surface Roughness*, *Notes*, *GD&Ts*, *Radii*, and *Threads*. A slight reduction is observed for *Title Blocks* in the category-specific model, where recall drops from 72.9% to 56.3%. This reduction is likely due to the variability and complexity of title block layouts, which are harder to learn with limited data in category-specific training.

The F1 score exhibits a similar trend, with the single model outperforming or matching the category-specific models in most feature categories. For example, F1 scores improve for *Surface Roughness* (94.6% vs. 79.1%), *Notes* (83.1% vs. 75.4%), *Title Blocks* (72.9% vs. 55.7%), and *Threads* (100% vs. 82.4%). F1 scores for *GD&T*, *General Tolerances*, and *Radii* remain consistent across both models. Lower F1 scores in *Materials* and *General Tolerances* are observed in both cases, primarily due to limited training samples.

The hallucination analysis further reinforces the precision advantage of the single model, which consistently exhibits lower hallucination across most categories. For example, the hallucination rate for *Surface Roughness* decreases from 34.6% (category specific) to 10.3% (single model), *Title Blocks* from 44.9% to 27.1%, and *Notes* from 39.5% to 28.9%. For *GD&T*, both models perform equally well, achieving a low hallucination rate of 5.23%, indicating strong and consistent performance. However, hallucination rates remain high for *Materials* and *General Tolerances* in both models. While the single model slightly reduces hallucination for *Materials* (from 83.3% to 80.0%), no improvement is observed in *General Tolerances*, where both models maintain a rate of 66.7%. These outcomes likely result from the scarcity of representative training data for these categories.

Overall, the single model demonstrates superior performance in precision, F1 score, and hallucination rate, while consistently maintaining high recall. Although the category-specific models achieve perfect recall in most cases, they exhibit reduced precision and limited generalization. These results indicate that the single model is better suited for broad, high-recall applications, whereas category-specific models may be advantageous only for precision-critical tasks involving well represented feature types.

## IV. DISCUSSION

The proposed hybrid framework, combining YOLOv11-obb for OBB detection with Donut for structured information parsing, demonstrates robust performance in extracting annotations from 2D engineering drawings. The single model configuration exhibits strong generalization, achieving perfect recall in most categories and significantly lower hallucination rates. This performance can be attributed to its exposure to the full annotation space, which enables it to capture a wide range of layout structures and annotation styles. These trends are clearly reflected in the evaluation metrics, where the single model consistently outperforms category-specific models in precision, F1 score, and hallucination rate across all feature categories.

In contrast, category-specific models, although designed for specialization, exhibit higher hallucination rates and lower precision in underrepresented categories such as *Materials* and *Surface Roughness*. Limited training data and reduced contextual diversity likely contribute to their lower generalizability and higher false positive rates.

A key advantage of the proposed framework is its ability to generate structured JSON outputs that are directly usable in downstream applications such as CAD/CAM systems, quality assurance tools, and digital manufacturing workflows. This contrasts with generic OCR systems, which often produce fragmented or unstructured outputs requiring additional post-processing. Furthermore, the framework supports semi-automated dataset labeling, reducing manual effort and improving scalability for future model training.

Beyond its performance, the modular design of the framework makes it adaptable to evolving manufacturing requirements. It can be extended to accommodate additional annotation categories and integrated into various stages of digital manufacturing pipelines. However, future work should focus on improving precision in underrepresented categories and reducing hallucinations in low quality drawings. These improvements will facilitate deployment at the industrial level.

## V. CONCLUSION

This paper proposed a hybrid deep learning framework for structured information extraction from 2D engineering drawings, integrating YOLOv11-obb for OBB detection and Donut for content parsing. The framework was evaluated using both single model and category-specific training strategies across four key performance metrics: precision, recall, F1 score, and hallucination rate. The single model consistently outperformed the category-specific models, demonstrating higher overall accuracy, stronger generalization, and fewer false positives.

The key contributions of this study included the development of a multi-category annotated dataset, the generation of structured JSON outputs compatible with downstream manufacturing systems, and a semi-automated labeling process that reduced annotation effort. Compared with traditional OCR based approaches, the proposed framework offers improved accuracy and greater integration potential for digital manufacturing workflows.

Future work will focus on expanding the dataset to include additional 2D drawing types and annotation styles, exploring multi-task training strategies, and evaluating alternative open-source vision-language models to enhance structured parsing in industrial applications. The framework shows strong potential to support digital manufacturing tasks such as process planning, quality control, and design verification in precision-focused industries including aerospace, automotive, and semiconductor manufacturing.

## ACKNOWLEDGMENT

This work is supported by the Singapore International Graduate Award (SINGA) (Awardee: *Muhammad Tayyab Khan*) funded by A*STAR, an AcRF Tier 1 grant (RG70/23) from Ministry of Education, Singapore, and Nanyang Technological University, and Korea Planning & Evaluation Institute of Industrial Technology (KEIT) grant funded by the Korea government (MOTIE) (No. RS-2024-00442448).